%% file: main.tex
\documentclass[conference]{IEEEtran}
\usepackage{times}
\usepackage[numbers]{natbib}
\usepackage{multicol}
\usepackage{amssymb}
\usepackage{amsmath}
\usepackage[bookmarks=true]{hyperref}
\usepackage{xcolor}
\usepackage{tabularx}
\usepackage{hyperref}       
\usepackage{url}            
\usepackage{booktabs}       
\usepackage{amsfonts}       
\usepackage{nicefrac}       
\usepackage{microtype}      
\usepackage{xcolor}         
\usepackage{lipsum}         
\usepackage{amsmath}
\usepackage{graphicx}
\usepackage{caption}
\usepackage{lipsum}
\usepackage{stfloats}
\usepackage{subfigure}
\usepackage[linesnumbered,ruled,vlined]{algorithm2e}

\newcommand{\ms}[2]{\small $\mathop{#1}_{\pm #2}$}

\makeatletter
\def\blfootnote{\xdef\@thefnmark{}\@footnotetext}
\makeatother

\begin{document}

\title{Offline Imitation Learning Through \\ Graph Search and Retrieval}

\author{Zhao-Heng Yin and Pieter Abbeel \\
UC Berkeley EECS
}

\twocolumn[{%
\renewcommand\twocolumn[1][]{#1}%

\maketitle
\begin{center}
    \includegraphics[width=0.99\linewidth]{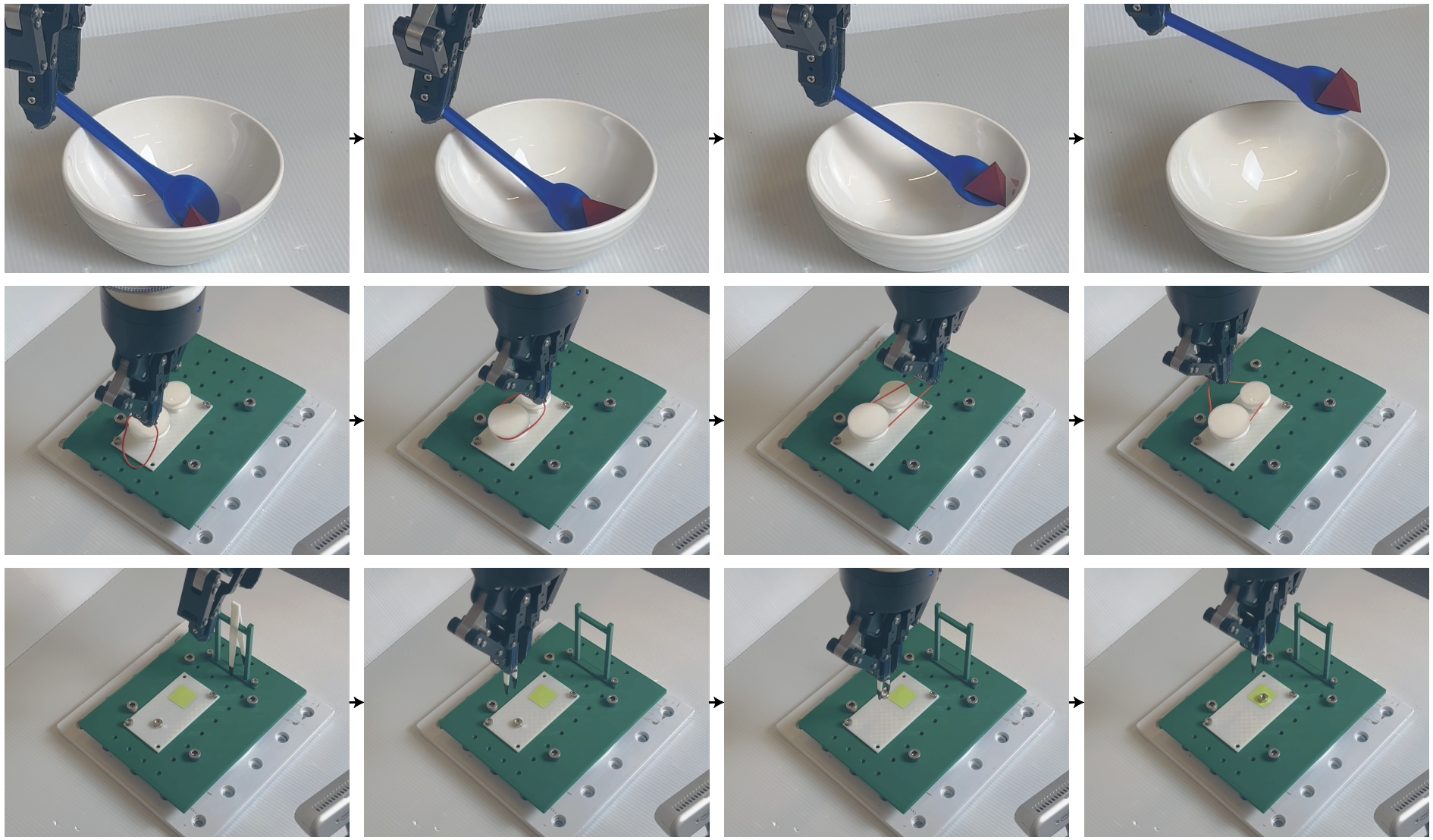}

    \captionof{figure}{Collecting high-quality human demonstrations for imitation learning can be very difficult. Consider the problem of using a spoon, tying a rubber band, and picking up tiny objects with a tweezer. These tasks require very precise and accurate manipulator movement. A human operator can exhibit suboptimal behaviors and retry multiple times during a single demonstration. We propose a novel algorithm called GSR to learn proficient behavior from such suboptimal data through graph search and retrieval.}
    \label{fig:teaser}
   \vspace{-0.1cm}
  
\end{center}
  }]

\begin{abstract}
Imitation learning is a powerful machine learning algorithm for a robot to acquire manipulation skills. Nevertheless, many real-world manipulation tasks involve precise and dexterous robot-object interactions, which make it difficult for humans to collect high-quality expert demonstrations. As a result, a robot has to learn skills from suboptimal demonstrations and unstructured interactions, which remains a key challenge. Existing works typically use offline deep reinforcement learning~(RL) to solve this challenge, but in practice these algorithms are unstable and fragile due to the deadly triad issue. To overcome this problem, we propose GSR, a simple yet effective algorithm that learns from suboptimal demonstrations through \underline{G}raph \underline{S}earch and \underline{R}etrieval. We first use pretrained representation to organize the interaction experience into a graph and perform a graph search to calculate the values of different behaviors. Then, we apply a retrieval-based procedure to identify the best behavior~(actions) on each state and use behavior cloning to learn that behavior. We evaluate our method in both simulation and real-world robotic manipulation tasks with complex visual inputs, covering various precise and dexterous manipulation skills with objects of different physical properties. GSR can achieve a 10\% to 30\% higher success rate and over 30\% higher proficiency compared to baselines. Our project page is at {\color{blue}{\url{https://zhaohengyin.github.io/gsr}}}.

\end{abstract}

\IEEEpeerreviewmaketitle

\input{src/1_introduction.tex}

\input{src/2_related_work.tex}
\input{src/3_background.tex}
\input{src/4_learning.tex}
\input{src/5_experiments.tex}

\input{src/6_conclusion.tex}

\bibliographystyle{plainnat}
\bibliography{references}
\newpage
\ \\ \\
\newpage
\input{src/appendix.tex}
\end{document}

%% file: src/1_introduction.tex
\section{Introduction}
Imitation learning~\cite{pomerleau1988alvinn, hussein2017imitation} is a powerful approach to learning robots that has achieved great success in robotic manipulation in recent years~\cite{shridhar2023perceiver, brohan2022rt, brohan2023rt, zhao2023learning, chi2023diffusion}. It involves training robots to mimic human demonstrations, allowing them to acquire manipulation skills in complex environments. Despite the remarkable strides made so far, we notice that most existing works usually assume expert-level task demonstrations, while many real-world robotic manipulation tasks involve precise, dexterous manipulation with complex objects which are hard for humans to provide high-quality demonstrations~(Figure \ref{fig:teaser}). For example, consider the task of making a robot manipulator grasp a tweezer to manipulate a tiny object, which is a common application scenario in a science laboratory. This task involves very fine-grained, precise control of gripping force and gripper movement. A human teleoperator can easily fail to provide successful demonstrations. Moreover, there usually exist suboptimal behaviors within a successful demonstration, such as retrying to grip the item if the first attempt fails. Naively using Behavior Cloning~(BC) to imitate the collected suboptimal demonstration can result in incapable control policies that can reproduce failure behavior~\cite{levine2020offline}.

We need an algorithm that can learn from suboptimal human demonstrations to solve this problem. Typically, a suboptimal human demonstration is composed of multiple segments of different expertise levels, and the whole demonstration dataset contains a set of useful temporal segments that combined together can approximate the behavior of a good policy. The desired learning algorithm should be able to pick up those good segments and recover this desired policy, which is also known as \textit{trajectory stitching} in reinforcement learning~(RL) research~\cite{levine2020offline}. In principle, it can be achieved by an offline deep-RL algorithm. However, in the application, previous research suggests that existing offline RL algorithms can usually struggle in realistic, long-horizon, pixel-based tasks~\cite{mandlekar2021matters, sutton2018reinforcement}. The simplest imitation learning algorithm like BC still performs the best among offline policy learning algorithms in many scenarios~\cite{mandlekar2021matters} and continues to be the standard practice in recent works.

To overcome the learning difficulties associated with typical offline RL, we propose GSR, a simple yet effective algorithm that learns from suboptimal demonstrations through \underline{G}raph \underline{S}earch and \underline{R}etrieval. Our key insight is to identify and stitch good behaviors directly through the use of graph representation and retrieval without boostrapping deep RL networks. Specifically, we leverage pretrained representations to build a graph representation for the collected interaction data. This graph serves as a nonparametric world model, with which we can reason about whether a temporal segment is proficient and can lead to desired future outcomes effectively through a graph search. Then, for each state in the dataset, we can retrieve the most proficient behavior on that state and use it for imitation learning~(Figure~\ref{fig:overview}). Our algorithm can be viewed as a filtering procedure to preprocess the dataset for imitation learning and can be used with any kind of deep imitation learning models~(architectures). As a direct approach that uses graph search rather than deep RL, our method enjoys high time efficiency. The whole preprocessing procedure typically consumes 10 to 30 minutes~(depend on the dataset) overhead compared to BC, making it practical for real-world applications.

In the experiments, we test our method in both simulation and real-world robotic manipulation tasks of various visual and physical complexities, involving high-resolution, multiview camera observation, and precise robot-object interactions. Despite numerous challenges in both perception and action, our method can consistently improve baselines' success rate by 10\% to 30\% and proficiency by over 30\%. We also provide various quantitative and qualitative analyses to show that our method is capable of identifying good behaviors in the dataset.

%% file: src/2_related_work.tex
\section{Related Work} \label{relatedworks}

\noindent \textbf{Learning from Suboptimal and Unstructured Data}
How to extract and learn good behaviors from suboptimal datasets is a central challenge in policy learning. When reward labels are available~(e.g. task completion reward in goal-conditioned setup in this paper), offline deep Reinforcement Learning~\cite{kumar2020conservative, levine2020offline, fujimoto2021minimalist, chen2021decision, janner2021offline, kostrikov2021offline, nair2020awac, wu2019behavior, kumar2019stabilizing, peng2019advantage} is one prevailing approach to address this problem. Most of the existing offline deep RL methods take the form of value-based learning. They learn value functions to estimate the goodness of actions with recursive Bellman updates and train policy to select good actions that lead to high values. However, deep RL is highly unstable and fragile due to deadly triad~\cite{zhang2021breaking, sutton2018reinforcement}, and people usually find that such (offline) RL algorithms struggle in application scenarios involving high-dimensional visual inputs and complex physical dynamics, and fail to outperform simple imitation learning algorithms like BC~\cite{mandlekar2021matters}. Our method is also based on value-based learning, but different from the existing methods, we directly use graph search and retrieval instead of prevailing deep RL updates to evaluate the value functions and perform policy optimization. This direct approach sidesteps the difficulty of bootstrapping deep value functions in complex environments~\cite{sutton2018reinforcement}.

When reward labels are not available, another line of work focuses on identifying proficient behaviors through other forms of feedback such as optimality labels, ranking, and preference~\cite{burchfiel2016distance, brown2019extrapolating, choi2019robust, xu2022discriminator,  chen2021learning, zhang2021confidence, beliaev2022imitation, myers2022learning, zhang2023discriminator, hejna2023few, kuhar2023learning}. This is an orthogonal research direction to the offline RL and our method. We believe our method can also benefit from these approaches by incorporating this additional information into the graph structure. This paper mainly considers a goal-conditioned setup and the proficiency can be defined through graph distances. 

\noindent \textbf{Graph and Retrieval in Policy Learning} 
Our method is built upon graph search and retrieval, and there are many policy learning works incorporating these techniques. Some works propose to incorporate a graph into the policy for goal-conditioned problems~\cite{savinov2018semi, eysenbach2019search, emmons2020sparse, zhang2021world, zhu2022value}. During inference, these methods involve locating the agent on the generated graph and running a graph search to decide the next subgoal. In other words, the graph can be considered as a high-level map-based planner to guide a low-level goal-conditioned policy. Our method differs from these works in the following ways. First, we only use the graph to estimate the goodness of behavior in the offline phase and do not use it as a high-level planner during inference which can fail to generalize and act robustly~(pick up inappropriate next goal in deployment)~\cite{emmons2020sparse}. Besides, we relate graphs to extracting better policies in an offline robotic dataset. Finally, we study the use of pretrained representations in building the graph in application scenarios for the first time, while existing works build the graph by training a specialized reachability network on the collected data~\cite{savinov2018semi}. 

Recently, people have found retrieval a useful tool for policy learning, especially in the field of robotics. For example, VINN~\cite{pari2021surprising} finds that a K-Nearest Neighbor policy with BYOL representation~\cite{grill2020bootstrap} makes a power robotic manipulation policy. Lately, Palo et al.~\cite{di2024effectiveness} use a retrieval process to find and execute relevant skill segments in the collected dataset to solve a robotic task. These works consider retrieval as part of the learned BC policy, while our method uses retrieval for sample weighting in offline policy learning. Perhaps the most related approach is Behavior Retrieval~\cite{du2023behavior}, which uses retrieval to find task-relevant transitions in a task-agnostic play dataset for imitation learning. We step further by using retrieval as a policy improvement operator, which is not studied by their work. 

%% file: src/3_background.tex
\begin{figure*}[t]
    \centering
    \includegraphics[width=\linewidth]{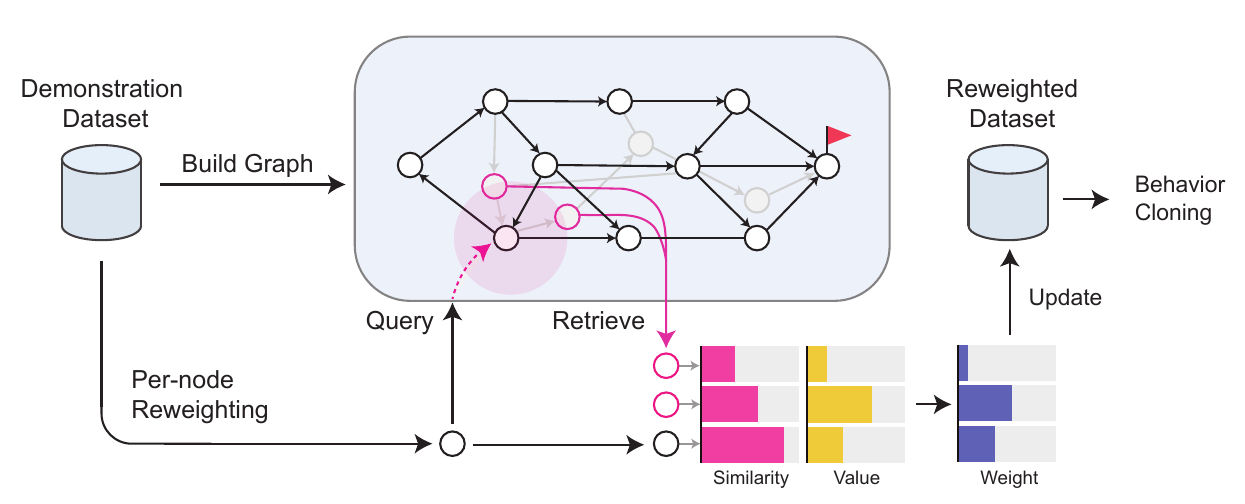}
    \caption{Overview of our algorithm. We first build a graph to represent the demonstration dataset and run a graph search to evaluate the goodness~(value) of each node. Then, we use a retrieval process to reassign weights to the transitions associated with each node. The weight reassignment is based on the similarity and goodness~(value) score. The entire process will give more weight to the optimal behaviors around each node, resulting in a reweighted dataset. We use this reweighted dataset for behavior cloning.}
    \label{fig:overview}
    \vspace{-0.2cm}
\end{figure*}
\section{Preliminaries}

\subsection{Problem Formulation}
In this paper, we study an offline policy learning setup. We provide the robot agent with a human interaction dataset $\mathcal{D}$ containing several task-relevant interaction trajectories. Each trajectory $\tau$ is a sequence of observations $o_{0:T}$ and corresponding actions $a_{0:T}$, i.e., $\tau=(o_0, a_0, o_1, a_1, ..., o_T, a_T)$. For each trajectory $\tau\in\mathcal{D}$, the human teleoperator controls the robot to solve the task and achieves success in the end, i.e., the last observation of the trajectory is a task success example. However, the demonstrated behavior can be highly suboptimal, including numerous errors and misses in the execution process. The setup favors the data collection process: It allows the data generated through human practice to serve as a demonstration, provided that the task is ultimately accomplished. We expect the robot to learn the task-solving skill by learning from $\mathcal{D}$ without using any online interaction samples.

\subsection{Offline Policy Learning}
Many offline policy learning algorithms usually involve training a policy network $\pi_\theta(a|o)$ to maximize the following weighted log-likelihood objective function:
\begin{equation}
    \mathcal{L}(\theta) = \mathbb{E}_{(o,a)\sim\mathcal{D}} [w(o,a)\log \pi_\theta(a|o)],
    \label{eqn:bc}
\end{equation}
where $w(o,a)$ is a predefined or learned weight function. For example, if we define $w(o,a)=1$ for all $(o,a)\sim\mathcal{D}$, the objective corresponds to the behavior cloning. If we define $w(o,a)=\exp(A(o, a))$ where $A$ is the advantage of taking action $a$ at observation $o$, this corresponds to the policy extraction objective used in Advantage-Weighted Regression~(AWR)~\cite{peng2019advantage} and Implicit Q-Learning~(IQL)~\cite{kostrikov2021offline}.

%% file: src/4_learning.tex
\section{Policy Learning \\by Graph Search and Retrieval}
GSR is a simple method that computes each weighting coefficient $w(o,a)$ using graph search and retrieval. Compared to the previous offline RL-based approach, it does not involve advantage computation with an extra Q-learning procedure, which is known to be unstable for high-dimensional pixel-based problems. For clarity, we first derive our idea step-by-step from a tabular case example in section~\ref{method:tabular}. We introduce the implementation details in the remaining sections.

\subsection{Overview}
\label{method:tabular}
For simplicity, let us first consider the problem of defining $w(o,a)$ in a tabular case where we want the agent to reach a certain goal observation $o_g$ efficiently at the fewest cost after learning. Here, the cost function is the number of steps taken to reach $o_g$. Obviously, for any observation $o$, the agent should pick an action $a$ whose resulting next observation~(more precisely, underlying state) $o_a$ has the shortest distance to the goal $o_g$. We denote the optimal action as 
\begin{equation}
    a^* = \arg\min_{a\in \mathcal{A}_o=\{a\in\mathcal{A}|(o,a)\in\mathcal{D}\}} d(o_a, o_g).
\end{equation}
The distance function $d$ is directly computable via a shortest path search. Then, we can define $w(o, a)=\mathbf{1}[a=a^*]$. In other words, we drop all the suboptimal goal-reaching actions from the dataset and only fit the remaining optimal ones. One may also use a softer version using $w(o,a)=\exp (-d(o_a, o_g))/Z$ to introduce stochasticity into the policy for future online fine-tuning, where $Z=\sum_{a\in{A_o}} \exp (-d(o_a, o_g))$ is a normalizer. Nevertheless, the above formula is not useful in practice. For the real-world continuous control problem, each encountered observation is unique in the human demonstration dataset $\mathcal{D}$, and we have the following two problems: (1) Similar states on different data trajectories are disconnected, and as a result, the distance estimate can be very inaccurate. (2) The argmin operator has only one candidate action ($|\mathcal{A}_{o_i}|= |\{a_i\}|=1$) and degenerates~($w(o_i,a_i)=1$ everywhere). In this case, no policy improvement can be performed. \ \\ \\
To address the first problem, we propose to identify and connect similar states in the dataset to form a better distance estimate in section~\ref{method:graph}. For the second problem, our insight is to create an approximated candidate set with nearest neighbor retrieval, and bias the weight towards relevant transitions that has higher value. We provide an overview of the proposed framework in Figure~\ref{fig:overview}.
\begin{figure}[t]
    \centering
    \includegraphics[width=0.95\linewidth]{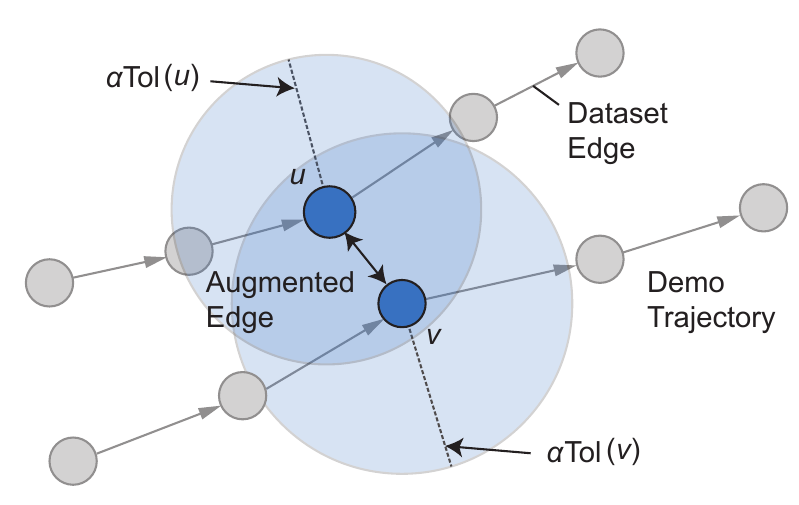}
    \caption{Identifying connectivity. Augmented edge: We add a bidirectional edge between two nodes $u$ and $v$ if they both lie in the tolerance range of each other in the pretrained representation space. Dataset edge: It represents ground truth transitions on each demonstration trajectory.}
    \label{fig:edge}
    \vspace{-0.3cm}
\end{figure}
\subsection{Graph Construction}
\label{method:graph}
The recent advance in large-scale representation pretraining provides us with powerful tools to identify similarities between different images. We can compare experiences from different trajectories in the representation space, and connect similar states to build a graph representation of the world. Concretely, we construct a graph $G(\mathcal{V}, \mathcal{E})$ to represent the whole dataset as follows.
\subsubsection{Building Vertex Set $\mathcal{V}$}
We define the vertex set $\mathcal{V}$ by subsampling each trajectory, resulting in a subset of the dataset observations. Specifically, for each trajectory $o_{0:T}$, we add $o_0, o_n, o_{2n}, ...$ to the vertex set, and we also refer to them as $v_0, v_1, ...$. The subsampling is simply used to make adjacent nodes on the data trajectory have enough visual difference~(adjacent nodes can look identical otherwise). To specify the task objective conveniently, we also add a ``virtual'' goal vertex $g$ to $\mathcal{V}$. 
\subsubsection{Building Edge Set $\mathcal{E}$} The \textit{directed} edge set $\mathcal{E}$ contains two kinds of edges: dataset edge and augmented edge. The dataset edge represents the transitions on each demo trajectory, serving as an approximation of the world dynamics. We connect each vertex to its next~(future) vertex on the same trajectory (i.e. $(v_0, v_1)$, $(v_1, v_2)$, ...). The last vertex of each trajectory corresponding to task success is connected to the goal vertex $g$. We set the weight of each dataset edge to 1.  \ \\ \\
So far, the vertices belonging to different trajectories are still disconnected from each other. We define augmented edges to bridge similar states across trajectories. To identify similar states, we use the off-shelf pretrained vision models to compute features for similarity computation. We first encode each vertex~(observation) $v_i$ with some pretrained encoder $f_\theta$, giving us the representation $f_{v_i} = f_\theta(v_i)$. Then, we determine a quantity called neighborhood tolerance for each node $v_i$ to decide connectivity, which we denote as $\rm {Tol} (v_i)$. We compute this using the following formulas:
\begin{align}
    &{\rm {Tol}_{raw}} (v_i) = \min \{{\rm sim}(f_{v_i}, f_{v_{i+1}}), {\rm sim}(f_{v_i}, f_{v_{i-1}})\},  \\
    &{\rm {Tol}} (v_i) = \frac{1}{M} \sum_{u\in \mathcal{N}_M(v_i)}{\rm {Tol}_{raw}} (u). 
\end{align}

\begin{algorithm}[t]
  \caption{GSR}
  [Optional] Finetune pretrained $f_\theta$ on $\mathcal{D}$\;
  Build graph $G(\mathcal{V}, \mathcal{E})$ using procedure in Section~\ref{method:graph}\;
  Set $w[v] = 0$ for each $v\in\mathcal{V}$\;
  Compute feature embedding $f_v$ for each $v\in\mathcal{V}$\;
   \For{$v\in\mathcal{V}$}{
        $\mathcal{N}(v)\leftarrow$ $\{v\}\smallskip \cup K-$Nearest neighbors of $v$\;
        \For{$u\in \mathcal{N}(v)$}{
            $w[u] \leftarrow w[u] + \frac{ \exp{[\beta_1 S(u, v) + \beta_2\tilde{Q}(u)]}}{\sum_{u'\in \mathcal{N}(v)} \exp{[\beta_1 S(u', v) + \beta_2\tilde{Q}(u')]}}$\;
        }
    }
  Set data weight $w(o,a)$ using $w$: $w(o_{in:(i+1)n}, a_{in:(i+1)n})\leftarrow w[v_i]$ for each $v_i\in\mathcal{V}$\;
  Train policy $\pi_\theta$ by maximizing Equation~\ref{eqn:bc}\;
  \label{alg:gsr}
\end{algorithm}

Here, ${\rm sim}$ is a metric to measure the similarity between features. In our case, it is negative $l_2$ distance ${\rm sim}(x,y)=-\Vert x - y\Vert_2$. In the first equation, we use the distance of each node to its one-step temporal neighbors as a rough neighborhood threshold. However, instead of using this single sample estimate, we average this using its $M$ nearest neighbors in the feature space, denoted as $\mathcal{N}_M(v_i)$. This nearest-neighbor set is computed using all the trajectories. Then, we connect two node $v_i$ and $v_j$ (bi-directionally) if 
\begin{equation}
    {\rm sim}(f_{v_i}, f_{v_j}) > \alpha \max\{{\rm Tol}(v_i), {\rm Tol}(v_j)\}.
\end{equation} 
Hence, we require each node to fall within each other's neighborhood range, multiplied by a hyperparameter $\alpha$. The procedure is shown in Figure~\ref{fig:edge}.  In practice, $\alpha$ is set to a value around 1: the connected vertices should be roughly within a 1-step reach from each other's view.

\subsection{Policy Improvement with Retrieval}
\label{method:retrieval}
With a constructed graph, we can calculate the value~(goodness) of each dataset transition from $v_i$ to $v_{i+1}$~(i.e. $(o_{in:(i+1)n}, a_{in:(i+1)n})$) using the distance on the graph. We define this as $\tilde{Q}(v_i) = Q(v_i,v_i\to v_{i+1}) =  -1 - d(v_{i+1}, g)$. With this function, we bias the action toward the better ones \textbf{for each} $v$. To do this, we first retrieve $K$ nearest neighbors of $v$ in the feature space, denoted $\mathcal{N}_K(v)$. Then, we reallocate the default dataset weight~(which is 1) assigned to $v$~(which is 1) to $v$ and these retrieved $K$ nearest neighbors. Intuitively, we should put high weight to a retrieved vertex $u$ only if (1) $u$ is similar to $v$ and (2) $u$ has a high value. More formally, inspired by the idea of locally-weighted regression~\cite{cleveland1988locally}, the weight allocated to each vertex $u\in \mathcal{N}(v) = \{v\}\cup \mathcal{N}_K(v)$ should be proportional to ${\rm Similarity}(u, v)\times \exp \tilde{Q}(u)$. Defining the first similarity term using an exponential form, we design the weight reallocation criterion at the vertex $v$ as
\begin{equation}
    w_{alloc}(u)|_v =  \frac{ \exp{[\beta_1 S(u, v) + \beta_2\tilde{Q}(u)]}}{\sum_{u'\in \mathcal{N}(v)} \exp{[\beta_1 S(u', v) + \beta_2\tilde{Q}(u')]}},
\end{equation}
where 
\begin{equation}
    S(u, v) = \frac{1}{2}\left(\frac{1}{|{\rm Tol} (u)|} + \frac{1}{|{\rm Tol} (v)|}\right) {\rm sim}(f_u, f_v)\leq 0
\end{equation}
is a normalized feature similarity function~(with respect to 1-step neighborhood distance computed at $u$ and $v$) to evaluate similarity between $u$ and $v$. $\beta_1$ and $\beta_2$ are two temperature hyperparameters, controlling the strength of the similarity term and value term, respectively. In particular, when $\beta_1\to \infty$ and $\beta_2\to 0$, the weight to distribute concentrates on $v$ and vanishes at the retrieved neighbors, and the policy learning reduces to the case of BC. Therefore, \textbf{our method smoothly interpolates between aggressive policy improvement and naive BC~(no policy improvement).}

\subsection{Implementation and Time Complexity}
\label{method:implementation}
We use R3M~\cite{nair2022r3m} as pretrained feature since it is pretrained with a contrastive objective, which we find can represent fine-grained movement well. In the task that involves multiple camera images, we calculate the feature embedding by concatenating the embedding of each image. We can also finetune the pretrained representation on the downstream task demo dataset~\cite{majumdar2024we} to make the retrieval more robust. We perform finetuning with a time-contrastive learning objective in simulated experiments and find that it improves the robustness to the selection of hyperparameters. We leave the discussion to the appendix. The pseudo-code of our method is summarized in Algorithm 1. The whole process is mainly composed of a graph search and single-round retrieval procedure, followed by BC. In particular, when we do not finetune the pretrained representation, the overhead computation compared to BC~(Line 2-9) in general takes less than 10 minutes in our experiments, making our method convenient to use. The time complexity of this procedure is $O(N\log N+|E|)$, where $N$ is the size of the dataset, and $|E|$ is the number of graph edges. Full implementation is in the appendix.

%% file: src/5_experiments.tex
\begin{figure}
    \centering
    \includegraphics[width=1\linewidth]{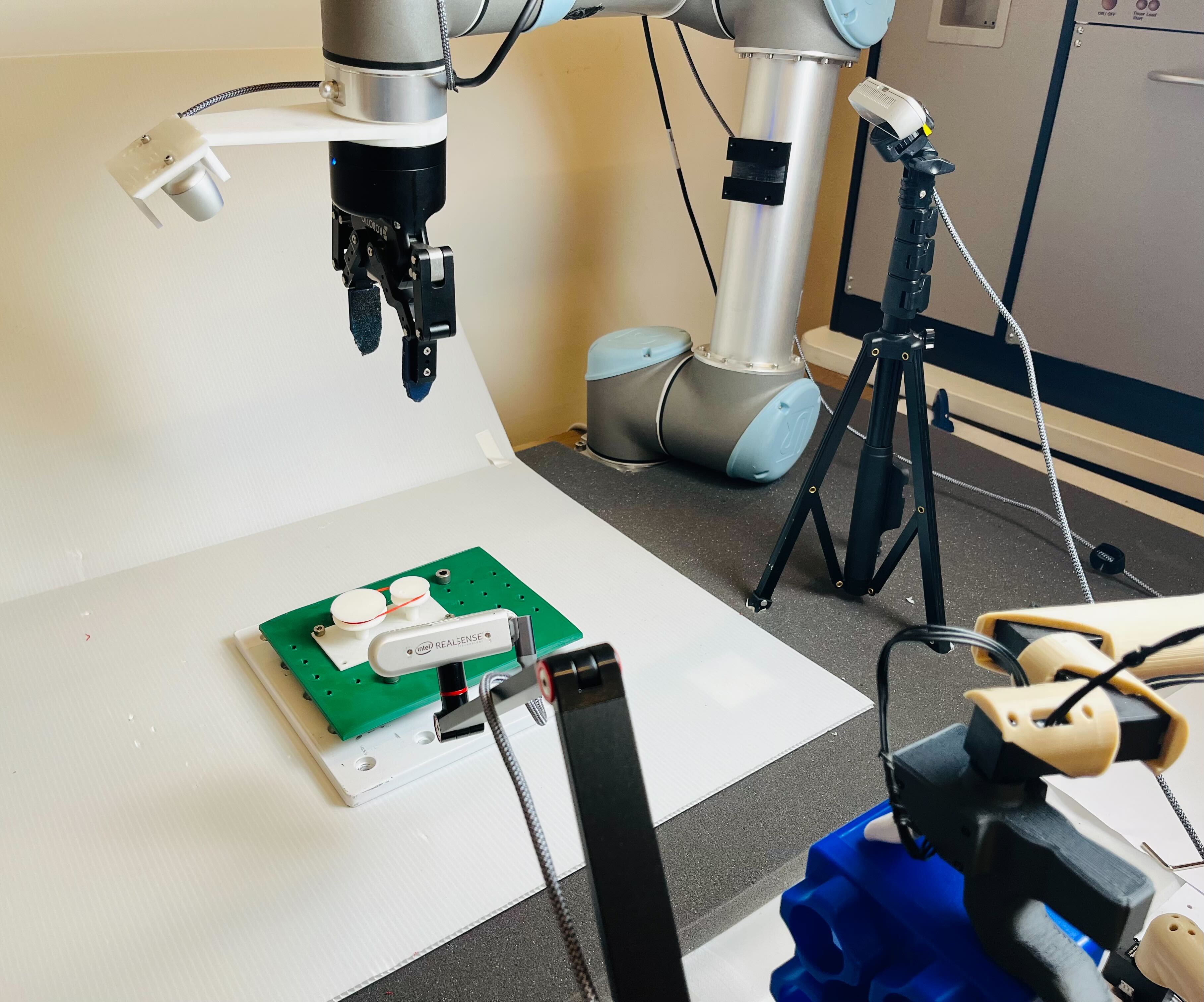}
    \caption{The real world robot manipulation setup. We conduct experiments on a UR5 robot arm with Robotiq gripper. We use 3 workspace cameras with $256 \times 256$ RGB observation, highlighting the challenge in perception especially for our considered precise manipulation tasks. }
    \label{fig:real_setup}
\end{figure}

\section{Experiments}
In experiments, we evaluate our method in both simulation and real world to validate its effectiveness in various robotic manipulation scenarios. We first study how much performance gain our method can achieve compared to the state-of-the-art imitation learning baseline. Then, through both quantitative and qualitative analysis, we show that our method identify and chain useful behaviors in the dataset to learn a robust policy. Finally, we present hyperparameter analysis to study design choices of our method.

\begin{figure}[t]
    \centering
    \includegraphics[width=0.95\linewidth]{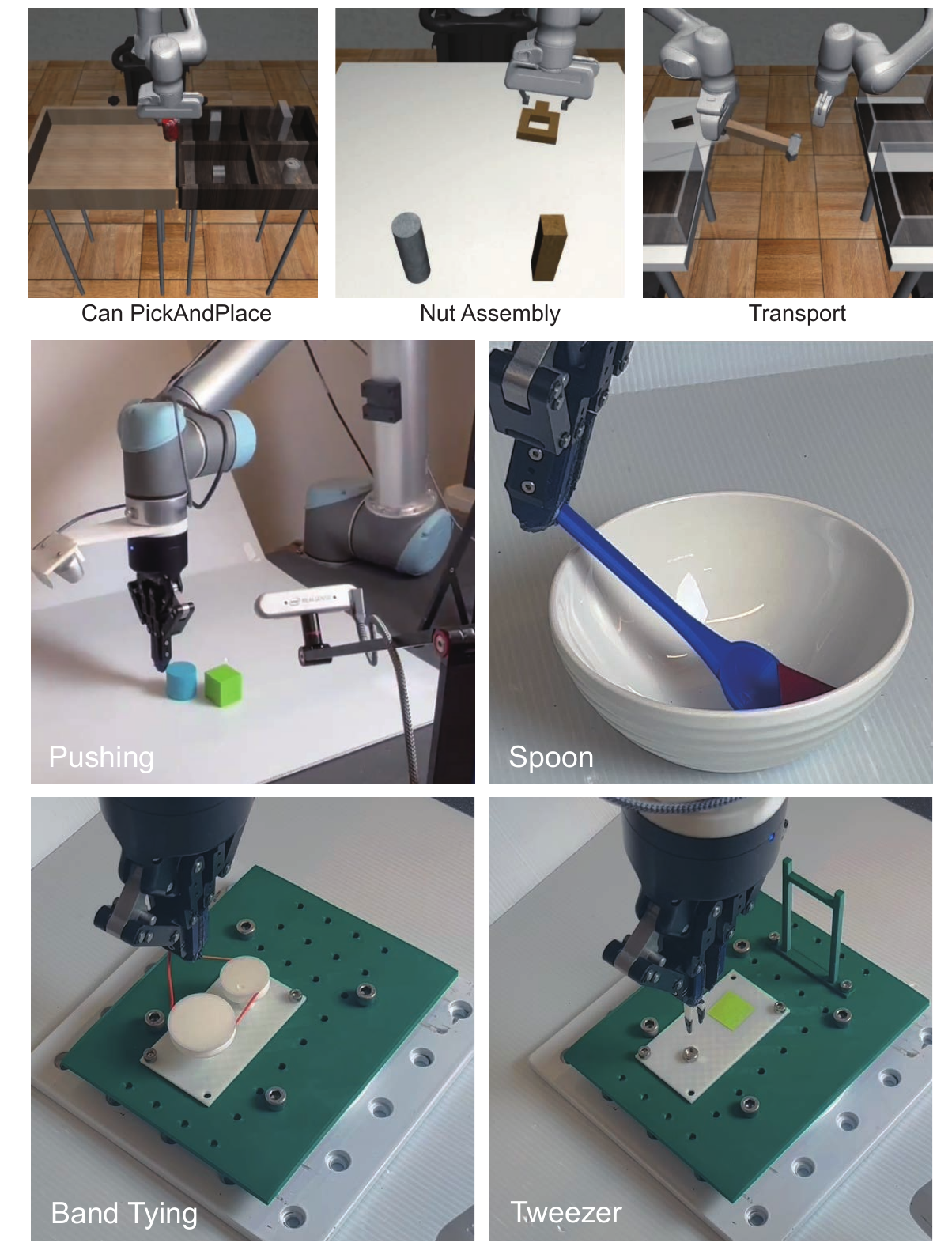}
    \caption{Illustration of the used tasks. Above: Tasks from the robomimic benchmark. Bottom: Our real-world tasks.}
    \label{fig:task_visualization}
\end{figure}
\subsection{Experiment Setup}
\subsubsection{Simulation Experiments}
We use the image-based Robomimic benchmark~\cite{mandlekar2021matters} as our testbed in simulation, which provides several robotic manipulation tasks with FrankaPanda robot. Specifically, we use three tasks that contain human demonstrations of diverse qualities~(Worse, Okay, Better):
\begin{itemize}
    \item \textbf{Can Pick-and-Place} In this task, the robot is required to pick up a can on the table and place it into a slot. We use the following two dataset setups: Worse-Okay20 and Worse-Better20~(the whole worse-human dataset with $20\%$ data of the better-human dataset).
    \item \textbf{Nut Assembly} In this task, the robot is required to pick up a square hole and plug it into a square peg. This task requires precise and accurate manipulation. We use the following two dataset setup: Worse-Okay50 and Worse-Better50.
    \item \textbf{Transport} This is a long-horizon dual arm manipulation task. One robot arm is required to pick up a hammer in a case and transport it to another arm. Then, the other arm should place the hammer on the target location. We use the default MH setup provided by the benchmark.
\end{itemize}
The number of Okay/Better demos is based on the difficulty of the task. The result of using different demos is in the appendix.
\subsubsection{Real-world Experiments}
In the real world, we design tasks of various complexities to highlight various aspects of our method. The robot workspace setup is shown in Figure~\ref{fig:real_setup}. 
\begin{itemize}
    \item \textbf{Pushing} This is the simplest task in our task suite. The robot is required to push a blue cylinder toward a green cube on the table. 
    \item \textbf{Spoon Scooping} In this task, the robot is required to pick up a spoon and use it to scoop a small object inside a slippery bowl. This task requires precise control of the spoon motion.
    \item \textbf{Tying a Rubber Band} This task is inspired by the NIST board challenge. The robot should pick up a thin rubber band and tie it onto a pair of wheels. This task highlights the challenge of robust perception against partial occlusion and fine-grained manipulation. 
    \item \textbf{Tweezer Manipulation} In this task, the robot needs to first pick up a tweezer that hangs on the shelf and then use the tweezer to pick up a nut and place it on a target platform. This task features both precise manipulation and long-horizon control. 
\end{itemize}
We use the GELLO system~\cite{wu2023gello} to collect human demonstrations, and the number of demonstrations and example rollouts can be found in the appendix. The Pushing and Spoon Scooping tasks use 2 cameras. The Rubber Band and Tweezer Manipulation tasks use 3 cameras. The camera images are all $256\times 256$ RGB images. The policy also uses the robot arm's proprioception information~(joint positions) as input. 

\subsubsection{Evaluation Metric}
To evaluate the performance of a trained policy, we use the following metrics. 
\begin{itemize}
    \item \textbf{Success rate~(SR)} is defined as the number of task successes divided by the number of trials in evaluation.
    \item \textbf{Time-to-success~(TTS)} is the average time a method takes to complete a task, which is computed only among the successful trials. A good method is expected to achieve a high success rate with low execution time. 
    \item \textbf{Normalized Proficiency~(NP)} is a metric we use for simulation. We calculate it by normalizing TTS with respect to the best human operator's mean TTS~(NP=1.0) and the worst human operator's mean TTS~(NP=0.0).
\end{itemize}

\input{src/table_robomimic.tex}
\input{src/table_real.tex}

\subsection{Baselines}
In the experiments, we mainly compare our method to the following baselines:
\begin{itemize}
    \item \textbf{Diffusion Policy~(DP)}. DP~\cite{chi2023diffusion} is a state-of-the-art behavior cloning baseline based on the diffusion model~\cite{ho2020denoising}. It is effective in overcoming the behavior multi-modality challenge in learning from human demonstrations in robotic manipulation. 
    \item \textbf{Implicit-Q-Learning~(IQL)}. IQL~\cite{kostrikov2021offline} is a strong offline RL baseline. We use R3M feature during Q-learning phase as in ~\cite{heo2023furniturebench} and implement its policy model as the diffusion policy above.
\end{itemize}
Our method also uses DP as the policy model. For the policy network architecture, we use separate ResNet-18~\cite{he2016deep} to process each input image, concatenate the output features with the proprioception information, and feed them into a Transformer-based denoiser network to generate the action. The whole policy network is trained from scratch. The denoising procedure is based on Denoising Diffusion Implicit Models~(DDIM)~\cite{song2020denoising} for fast deployment time inference. Specifically, in the simulation experiment, the last average pooling layer of ResNet-18 is replaced by a spatial softmax~\cite{finn2016deep} as in previous works~\cite{mandlekar2021matters, chi2023diffusion}. 

\subsection{Simulated Results}
Table~\ref{table:sim} shows our results in simulation. First, we observe that our method can reduce the average time to success across different tasks. Compared to previous methods, we can improve NP by 30\%-50\% on Can PickAndPlace and Nut Assembly, and can reach the best human operator's time efficiency in some cases. Note that the best human operator's proficiency is already high enough in these tasks, so we do not expect our algorithm's NP to exceed the human operator's performance~($>$100\%).  The second finding is that our algorithm can also improve the success rate of state-of-the-art algorithms. The effect is also task-dependent. Interestingly, the improvement is the most significant (around 14\%) in the most difficult long-horizon bimanual transport task, while for the other simpler tasks like NutAssembly, the improvement is milder~(5\% to 10\%). We conjecture that this is because all the used demonstrations eventually lead to successes and the suboptimal behaviors in those datasets are mainly caused by slow motion or minor corrections. In this case, following the suboptimal trajectories will not harm the success rate much. 

\begin{figure}[t]
    \centering
    \includegraphics[width=1\linewidth]{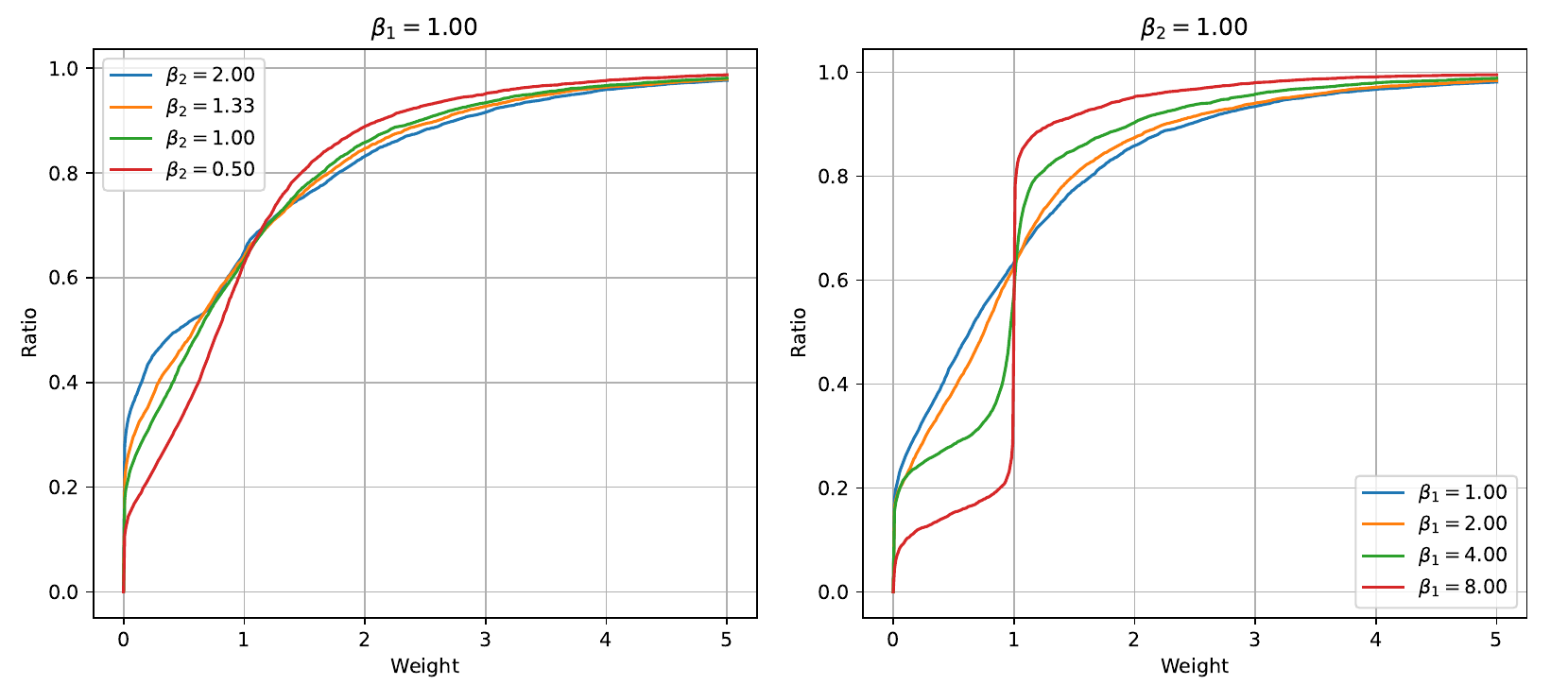}
    \caption{Dataset weight distribution plot (cumulative distribution function using sample statistics) v.s. different temperatures. As $\beta_1$ decreases or $\beta_2$ increases, we can see a larger contrast within samples~(more samples away from weight 1). The plot is based on Can-WB20.}
    \label{fig:weight-distribution}
    \vspace{-0.3cm}
\end{figure}

\begin{figure}[ht]
    \centering
    \includegraphics[width=\linewidth]{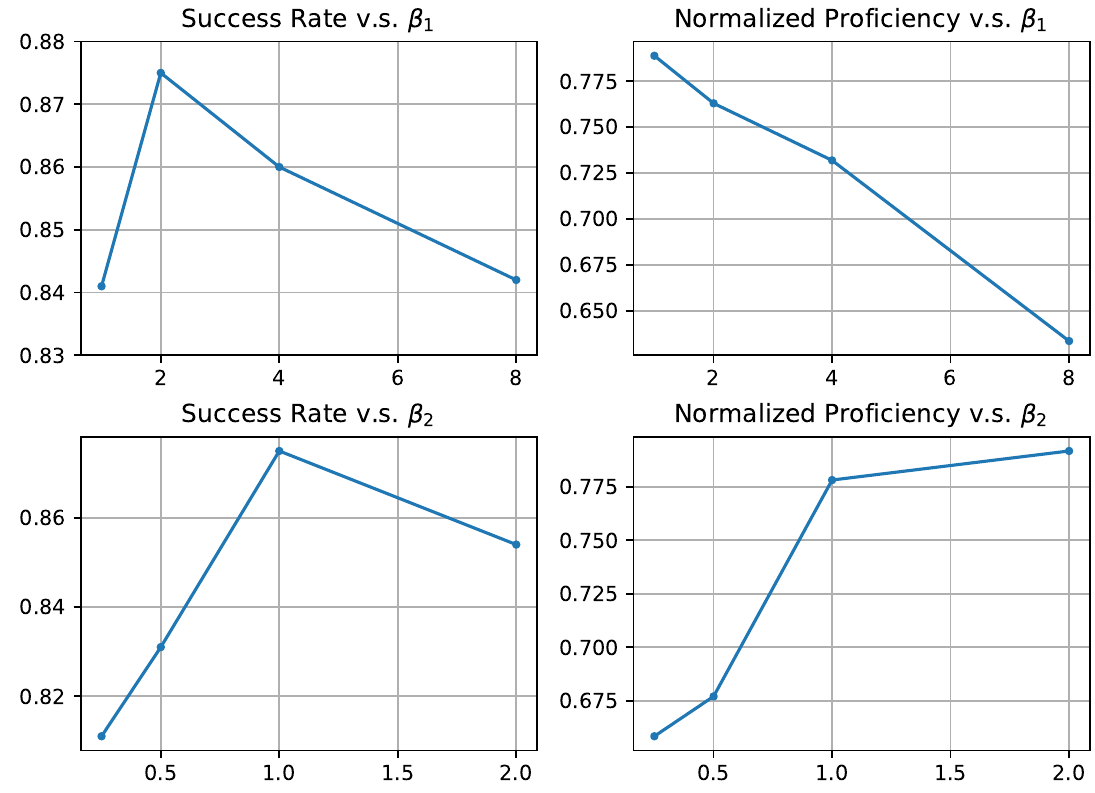}
    \caption{Success rate and Normalized Proficiency metric under different $\beta_1$ and $\beta_2$. The results are computed on the Can WB20. In general, we observe that decreasing $\beta_1$ and increasing $\beta_2$ can lead to improved success rate and proficiency for a wide range of $\beta$. }
    \label{fig:ablation}
\end{figure}

\begin{figure*}[ht]
\centering
\includegraphics[width=\linewidth]{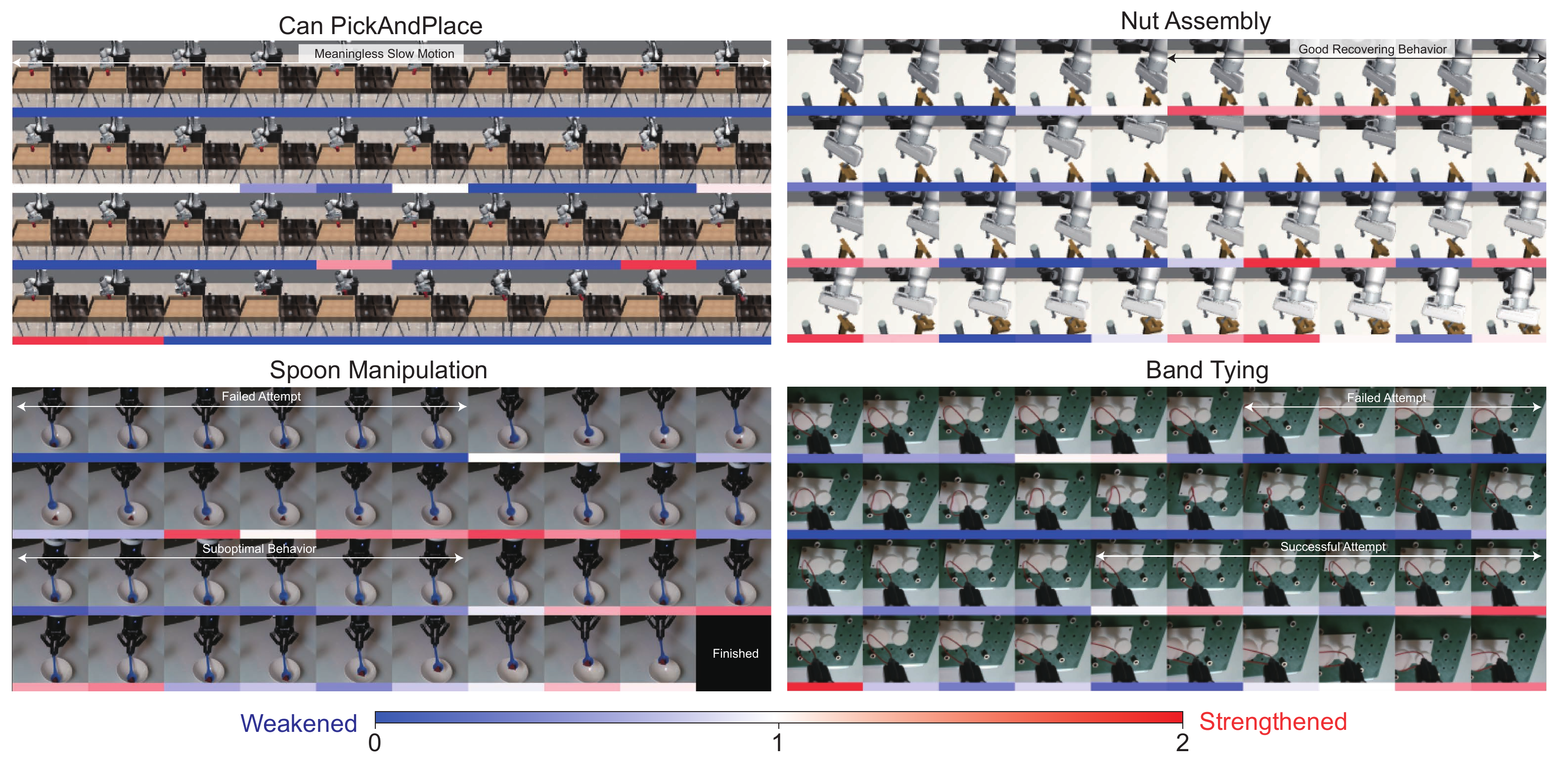}
 \caption{Visualization of calculated weight $w$ for transitions on a demonstration trajectory. We select 4 trajectories~(part) from 4 different tasks. We use red to indicate strengthened transitions and blue to indicate weakened transitions. Frame Order: Left to Right, Top to Bottom.}
\label{fig:qual}
\vspace{-0.4cm}
\end{figure*}

\begin{figure}[ht]
    \centering
    \includegraphics[width=\linewidth]{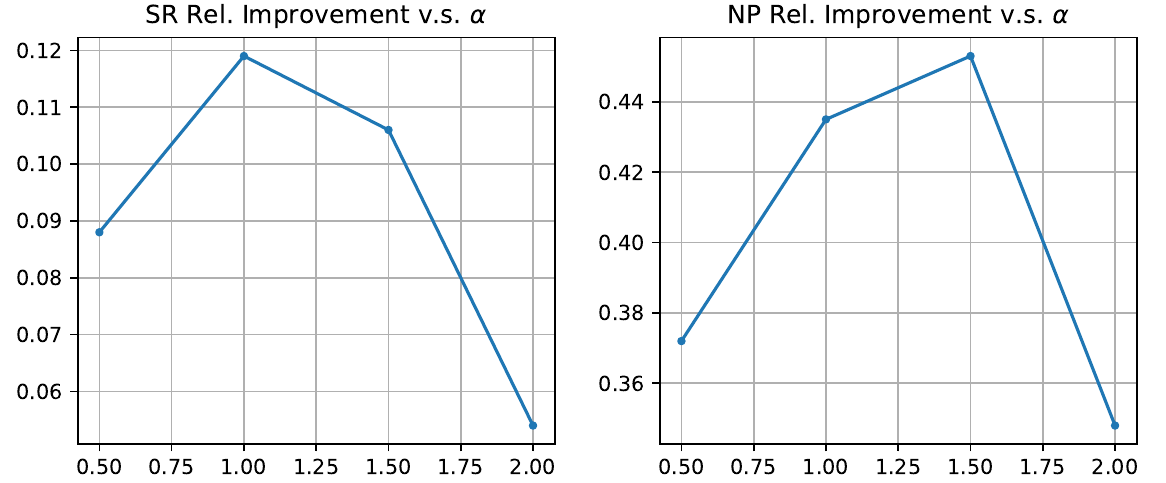}
    \caption{Relative Success rate and Normalized Proficiency improvement compared to baseline BC~(DP) under different $\alpha$. The results are averaged on the Can and NutAssembly.}
    \label{fig:alpha_ablation}
    \vspace{-0.5cm}
\end{figure}

\subsection{Real-world Results}
As shown in Table~\ref{table:real}, our results in simulation also transferred to the real-world experiments well. We find that our method can achieve a success rate greater than 80\% in the considered task and outperform all baselines in execution time. We observe that the baselines appear to repeat the failed human attempts during demo collection more frequently compared to our method. In some cases, they may eventually recover from failed behavior modes (by repeating subsequent retrying behavior demonstrated by a human) and complete the task. However, in many cases, they will get stuck or go out of distribution, leading to a complete failure. In contrast, our method can identify the good behavior in the dataset, and avoid learning these failure behavior patterns~(see next sections).

\subsection{Qualitative Results: What is our method learning?}
To understand what kind of behavior our policy is trying to learn, we perform qualitative analysis by visualizing our weight function $w$. We conduct the study in both simulation and real-world experiments. The result is shown in Figure~\ref{fig:qual}. We use red and blue colors to indicate high~(strengthened behavior) and low~(weakened behavior) $w$ respectively. Interestingly, we have the following findings: (1) All the temporal segments that lead to the failures are weakened and have low weights. (2). However, the segments that demonstrate failure-recovering behavior are not weakened because the best option for these states is to follow the dataset transitions. (3). The proficient temporal segments are usually strengthened as desired, and the segments that correspond to slow movement are slightly weakened. We provide more visualization results in the appendix. 

\subsection{Hyperparameter Analysis}
Having known that our method indeed strengthened desired behavior, in this section, we further study the effect of the main hyperparameters in our algorithm:
\paragraph{$\beta_1$ and $\beta_2$} These temperatures control the policy improvement strength as discussed in section~\ref{method:retrieval},. We would like to understand how they affect weight distribution over the dataset and the resulting learning performance. We visualize the weight distribution in Figure~\ref{fig:weight-distribution}. We can see that as $\beta_1$ decreases or $\beta_2$ increases, more data has lower weight and is rejected~(zero weight), and we can observe a stronger contrast in the weight distribution, with fewer samples' weight center around 1. Then we plot the change in success rate and normalized proficiency in Figure~\ref{fig:ablation}. We find that both success rate and proficiency increase as we decrease $\beta_1$ or increase $\beta_2$~($\beta_1$ decrease: using more dissimilar samples; $\beta_2$ increase: biasing more towards proficient samples) for a wide range of $\beta$s. However, when $\beta_1$ becomes too small or $\beta_2$ becomes too large, we find the success rate can drop. Intuitively, this is because less data is used for training. Although the policy can become more proficient in observed (narrower) regions, it may fail to solve the task once it deviates from the data distribution.
\paragraph{Tolerance $\alpha$} $\alpha$ controls the range of tolerance and decide connections. We sweep it over different values to see how it impacts learning outcome~(Figure~\ref{fig:alpha_ablation}). We find that both low and high $\alpha$ will affect the success rate and proficiency. A low alpha will kill most of the augmented edges, and can provide an inaccurate distance estimate. On the other hand, a high alpha can connect nodes that are actually far away from each other and also harm distance estimate. Therefore, we need to specify $\alpha$ in an appropriate range for best performance. We find that $\alpha$ can be tuned conveniently through some visualizations: we simply need to set $\alpha$ to different values examine whether the generated edges contain obvious errors~(such as connecting two very dissimilar states) and select a large $\alpha$ whose error ratio is below a certain threshold.

%% file: src/table_robomimic.tex
\begin{table*}
\centering
\captionof{table}{Performance of evaluated methods on the \textbf{image-based} Robomimic benchmark. The results are averaged on 3 different seeds. For each seed, the evaluation is averaged over the last 5 checkpoints in training. Each evaluation takes 50 trials. We provide sufficient time for each policy to succeed~(according to the recommendation of the benchmark). Our proposed method can outperform the baselines in both success rate~(SR) and normalized proficiency~(NP) across different tasks. } 
\normalsize
\renewcommand\arraystretch{1.1}
\setlength\tabcolsep{3pt}
 \begin{tabular*}{\textwidth}{l@{\extracolsep{\fill}}cccccccccc}
\toprule
\textbf{Task} &\multicolumn{2}{c}{\textbf{Can~(WO)}} & \multicolumn{2}{c}{\textbf{Can~(WB)}} & \multicolumn{2}{c}{\textbf{NutAssembly~(WO)}} & \multicolumn{2}{c}{\textbf{NutAssembly~(WB)}} & 
\multicolumn{2}{c}{\textbf{Transport}}  \\
& SR$(\uparrow)$ & NP$(\uparrow)$    & SR$(\uparrow)$ & NP$(\uparrow)$  & SR$(\uparrow)$ & NP$(\uparrow)$   & SR$(\uparrow)$ & NP$(\uparrow)$   & SR$(\uparrow)$ & NP$(\uparrow)$    \\
\hline

DP & \ms{0.77}{0.02} & \ms{0.46}{0.06} & \ms{0.80}{0.02} & \ms{0.47}{0.06}  & \ms{0.44}{0.03} & \ms {0.48}{0.05}  & \ms{0.48}{0.02} & \ms{0.62}{0.01}  & \ms{0.53}{0.02} & \ms {0.54}{0.02}    \\
IQL-DP & \ms{0.77}{0.01}  &  \ms{0.57}{0.04} & \ms{0.81}{0.01} & \ms{0.52}{0.03}  & \ms{0.40}{0.02} & \ms{0.48}{0.08}  & \ms{0.46}{0.02} & \ms{0.33}{0.03}  & \ms{0.53}{0.02} & \ms{0.56}{0.02}   \\
Ours & \ms{\textbf{0.85}}{\textbf{0.02}}  & \ms{\textbf{0.86}}{\textbf{0.02}}& \ms{\textbf{0.87}}{\textbf{0.01}} & \ms{\textbf{0.76}}{\textbf{0.03}}  & \ms{\textbf{0.53}}{\textbf{0.01}} & \ms{\textbf{1.16}}{\textbf{0.03}}  & \ms{\textbf{0.57}}{\textbf{0.04}} & \ms{\textbf{0.88}}{\textbf{0.03}}  & \ms{\textbf{0.67}}{\textbf{0.03}} & \ms{\textbf{0.63}}{\textbf{0.02}}   \\
\bottomrule
\end{tabular*}
\label{table:sim}
\end{table*}

%% file: src/table_real.tex
\begin{table*}

\centering
\captionof{table}{Performance of evaluated methods on the real-world tasks. The results are averaged on the last checkpoint of 3 seeds. Each checkpoint evaluation takes 30 trials.  We provide sufficient time for each policy to succeed. The time-to-success~(TTS) metric is measured in seconds.} 
\normalsize
\renewcommand\arraystretch{1.1}
\setlength\tabcolsep{3pt}
 \begin{tabular*}{\textwidth}{l@{\extracolsep{\fill}}cccccccc}
\toprule
\textbf{Task} &\multicolumn{2}{c}{\textbf{Pushing}} & \multicolumn{2}{c}{\textbf{Spoon Scooping}} & \multicolumn{2}{c}{\textbf{Band Tying}} & \multicolumn{2}{c}{\textbf{Tweezer}} \\
& SR$(\uparrow)$ & TTS$(\downarrow)$    & SR$(\uparrow)$ & TTS$(\downarrow)$   & SR$(\uparrow)$ & TTS$(\downarrow)$   & SR$(\uparrow)$ & TTS$(\downarrow)$       \\
\hline

DP & \ms{0.58}{0.14} & \ms{12.0}{1.6} & \ms{0.63}{0.08} & \ms{30.2}{1.8}  & \ms{0.57}{0.14} & \ms {34.8}{2.6}  & \ms{0.48}{0.04} & \ms{41.2}{2.8}     \\
IQL-DP & \ms{0.82}{0.07}  &  \ms{9.4}{0.8} & \ms{0.67}{0.08} & \ms{29.4}{1.3}  & \ms{0.57}{0.06} & \ms{30.6}{2.4}  & \ms{0.46}{0.12} & \ms{37.0}{2.0}    \\
Ours & \ms{\textbf{0.95}}{\textbf{0.03}}  & \ms{\textbf{7.8}}{\textbf{0.4}}& \ms{\textbf{0.81}}{\textbf{0.05}} & \ms{\textbf{24.0}}{\textbf{1.9}}  & \ms{\textbf{0.86}}{\textbf{0.05}} & \ms{\textbf{22.5}}{\textbf{1.4}}  & \ms{\textbf{0.76}}{\textbf{0.06}} & \ms{\textbf{29.7}}{\textbf{2.5}}     \\
\bottomrule
\end{tabular*}
\label{table:real}
\end{table*}

%% file: src/6_conclusion.tex
 \section{Conclusion}
 In this paper, we have presented GSR, a simple offline imitation learning method that can learn from suboptimal datasets through graph search and retrieval. We carried out experiments in both challenging simulation and real-world environments to demonstrate its effectiveness. In the future, there are several open problems to study. For example, how do we improve the quality of the representation to make our algorithm more powerful? In this paper, we considered demonstrations in the same context~(i.e., the same workspace). However, it would be more desirable to have a task-invariant representation that can bridge experience from different contexts~(backgrounds). Besides, in the context of large-scale pretraining, it would be interesting to use pretrained forward prediction models to generate more long-horizon connections, rather than a 1-step connection in this paper. This can leverage general knowledge from other relevant tasks for policy optimization.  We hope that our work can inspire more powerful policy optimization methods.

 \section*{Acknowledgement}
This work is supported by InnoHK Centre for Logistics Robotics and ONR MURI N00014-22-1-2773. The authors would like to thank Philipp Wu, Yide Shentu, and Toru Lin for the hardware support.

%% file: src/appendix.tex
\appendix

\subsection{Model and Training Details}
For both the simulated and real-world experiments, we use ResNet-18 as the feature extractor.  For the action denoiser, we follow the setup used in the Diffusion Policy paper~\cite{chi2023diffusion}. Specifically, we use a transformer-based action denoiser. We use an embedding dimension of 256, and 8 layers with 4 heads. We use AdamW optimizer to train our policy model. The learning rate is set to 0.0001.  For the simulated tasks, we train the policy for 3000 epochs. For the real-world tasks, we train the policy for 2000 epochs. For the DDIM noise scheduler, we use 100 training steps with a squared cosine $\beta$ schedule. We use 10 denoising steps during DDIM inference. We normalize both input observation and action into $[-1, 1]$. For the image input, we apply a random crop data augmentation during training as in previous works. We apply a fixed CenterCrop to the observed image during inference. The setup of the hyperparameters is shown in Table~\ref{tab:hp}.

\subsection{Representation Finetuning}
Although pretrained representation can serve a broad range of tasks, it may still require finetuning for better retrieval performance especially when the observation in the downstream task is not covered by the pretraining dataset. We provide a simple method for adapting the representation. We finetune the representation on the downstream task trajectories with a time contrastive loss used by R3M. Specifically, we freeze the pretrained network $f$ and introduce a trainable MLP adapter head $g$. Then, we train $g$ to maximize the following objective: 
\begin{align*}
    &\mathcal{L} = \\& \mathop{\mathbb{E}}\limits_{(x,\{x_n^i\},x_p)\sim \mathcal{D}} \log \left[\frac{\exp{(S(z, z_p) / \tau)}}{\sum_i \exp{(S(z, z_n^i) / \tau)} + \exp {(S(z, z_p) / \tau)}}\right],
\end{align*}
where $x$ is the anchor sample, $\{x_n^i\}$ is a set of negative samples, $x_p$ is the positive sample, and $z = g(f(x))$. $S(x,y) = -\Vert x-y\Vert^2$ is the L2 distance function. $\tau=0.1$ is a temperature hyperparameter. The selection of negative and positive samples is the same as that used in the R3M paper~\cite{nair2022r3m}. After training, we use $g(f(x))$ as the finetuned representation. We apply representation finetuning in the simulation experiments. $g$ is an MLP with hidden layer size [512, 256] which outputs a 128-d vector. We use a batch size of 128 and a learning rate of 0.0003. We find that without representation finetuning, our approach can be more sensitive to hyperparameters in simulation. Specifically, the success rate and normalized proficiency can drop by 4\% and 12\% respectively, on average, across different $\beta_1$ used in our hyperparameter analysis part. However, the performance of the best hyperparameter setup without finetuning can still match that of the finetuned approach.

Since the representation network $f$ is frozen in this procedure and the only trainable network is a shallow MLP $g$, the representation finetuning procedure is also computationally efficient.

\begin{table}[th]
\centering
\captionof{table}{Performance of different methods when using different number of demos.} 
\normalsize
\renewcommand\arraystretch{1.1}
\setlength\tabcolsep{3pt}
 \begin{tabular*}{\linewidth}{l@{\extracolsep{\fill}}cccc}
\toprule
\textbf{Task} &\multicolumn{2}{c}{\textbf{Can50~(WB)}} & \multicolumn{2}{c}{\textbf{NutAssembly20~(WB)}} \\
& SR$(\uparrow)$ & NP$(\uparrow)$    & SR$(\uparrow)$ & NP$(\uparrow)$ \\
\hline
DP & \ms{0.85}{0.02} & \ms{0.58}{0.03} & \ms{0.41}{0.02} & \ms{0.28}{0.04} \\
IQL-DP & \ms{0.87}{0.01}  &  \ms{0.63}{0.02} & \ms{0.38}{0.01} & \ms{0.25}{0.03} \\
Ours & \ms{\textbf{0.93}}{\textbf{0.01}}  & \ms{\textbf{0.82}}{\textbf{0.03}}& \ms{\textbf{0.47}}{\textbf{0.02}} & \ms{\textbf{0.51}}{\textbf{0.04}}  \\
\bottomrule
\end{tabular*}
\label{table:demo_perf}
\end{table}

\begin{table}[t!]
    \centering
    \normalsize
     \caption{Hyperparameter Setup for differnt tasks.}
    \begin{tabular}{cccc}
        \toprule
        Task & $\beta_1$ & $\beta_2$ &  $\alpha$ \\
        \midrule
        Can &   2.0 & 1.0 &  1.0 \\
        Nut Assembly &2.0  & 0.33 & 1.0 \\
        Transport& 2.0   & 0.25 & 1.0 \\
        \midrule
        Pushing &   1.0 &  0.25 &1.0 \\
        Spoon &     1.0 &  0.25 & 1.0 \\
        Band Tying &   1.0 & 0.25 & 1.25 \\
        Tweezer &    1.0 &  0.25 & 1.25 \\
        \bottomrule
    \end{tabular}
    \label{tab:hp}
\end{table}

\subsection{Human Demos}
We use 100 demonstrations for the real world tasks. We randomly initialize the object and the robot pose at the beginning of each demo collection. We also slightly randomize the camera pose in the workspace and the light during the demo collection process. Some human demo examples can be found in the trajectory visualization in section~\ref{appendix:qual}.

\subsection{Performance under Different Number of Demos}
We provide the performance of different approaches when using different numbers of demos in Can PickAndPlace and Nut Assembly in Table~\ref{table:demo_perf}. Our method can achieve similar performance gains in success rate and proficiency.

\subsection{More Qualitative Results}
\label{appendix:qual}
\begin{figure*}
    \centering
     \includegraphics[width=0.85\linewidth]{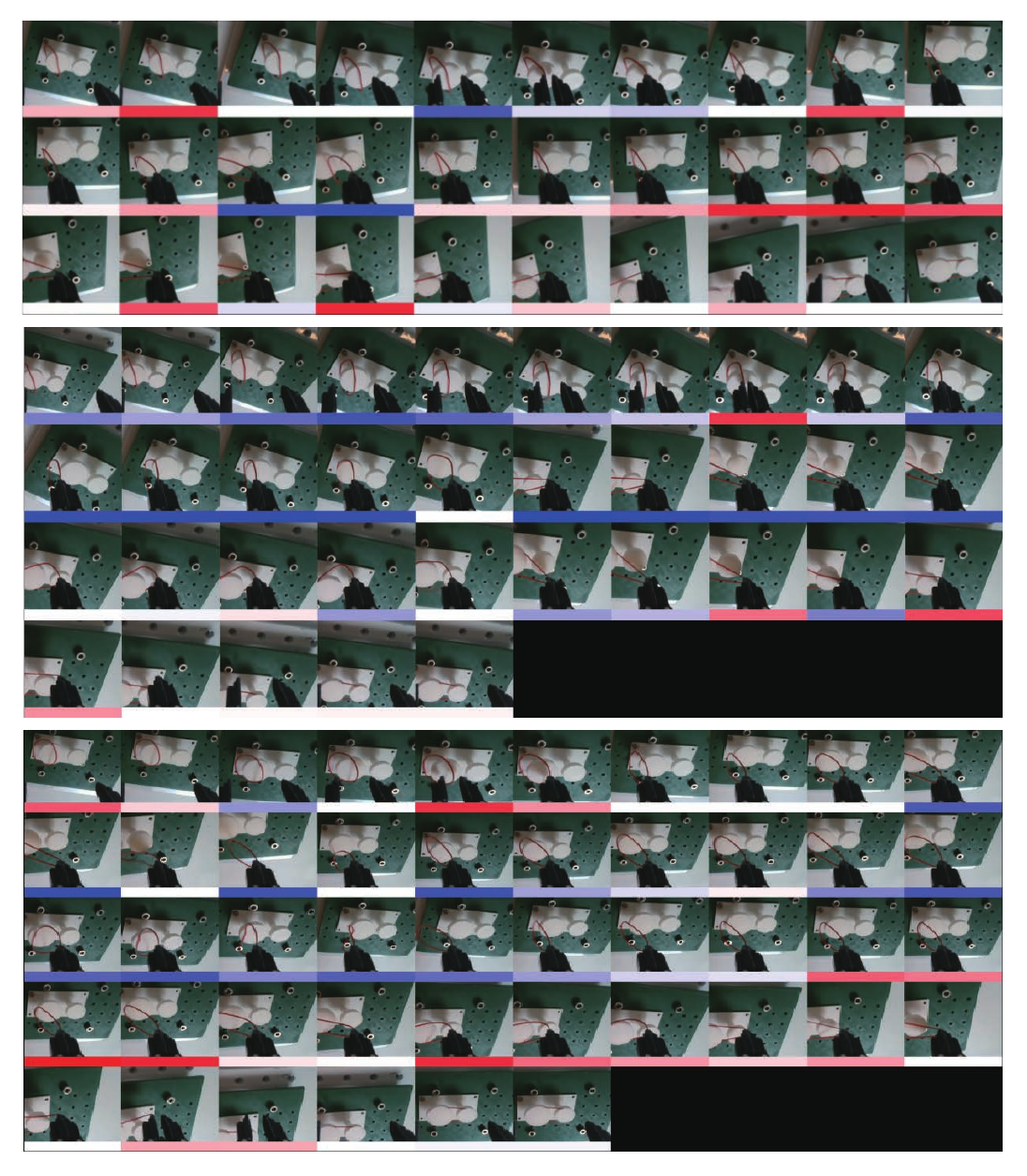}
    \caption{More Weight Visualization (Band Tying). }
    \label{fig:sup3}
\end{figure*}

\begin{figure*}
    \centering
     \includegraphics[width=0.85\linewidth]{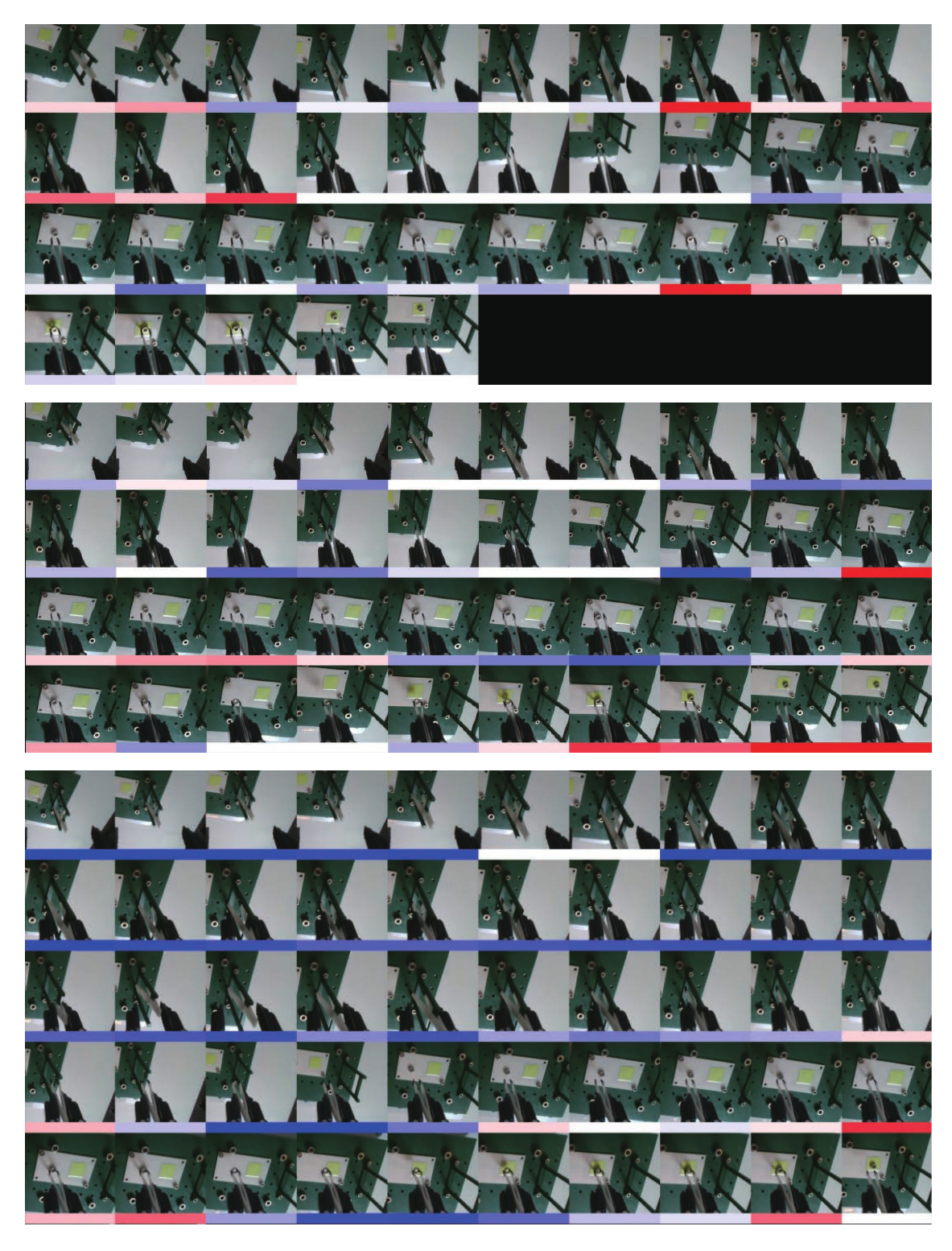}
    \caption{More Weight Visualization (Tweezer).}
    \label{fig:sup4}
\end{figure*}

\begin{figure*}
    \centering
    \includegraphics[width=0.85\linewidth]{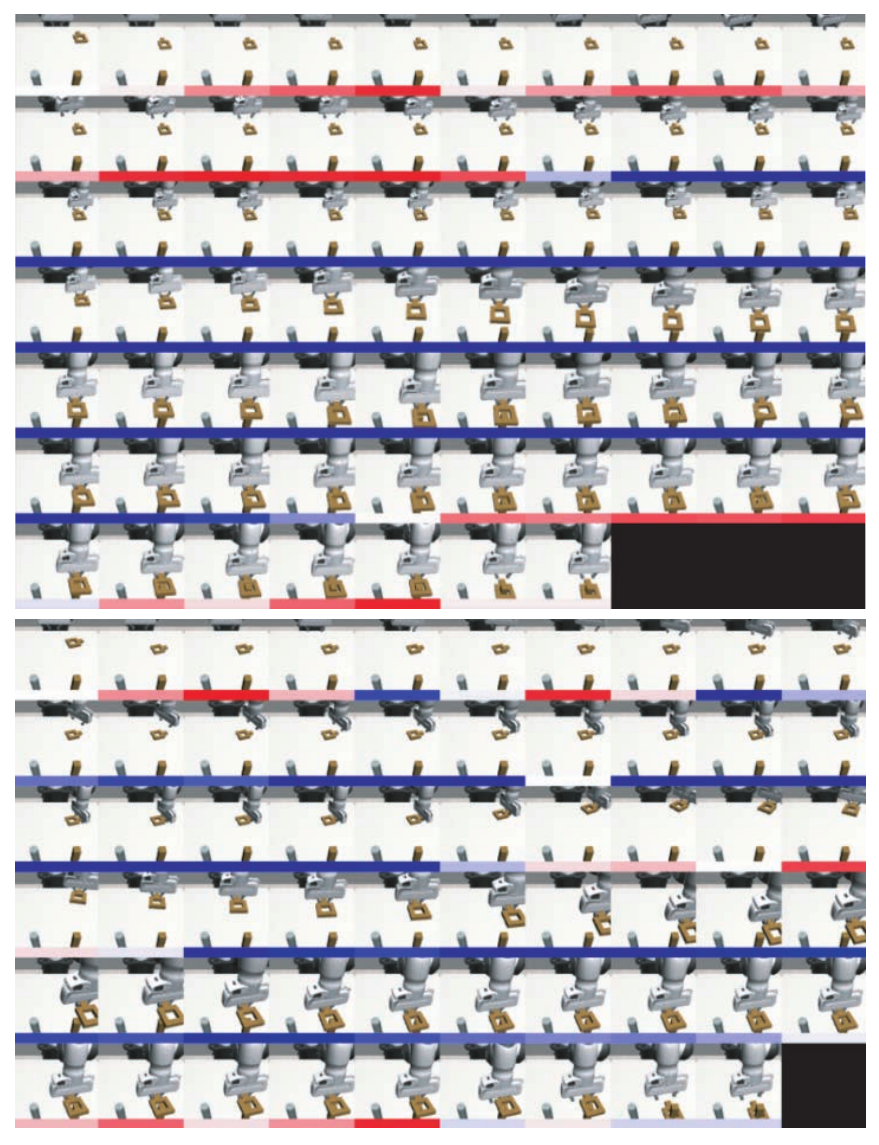}
    \caption{More Weight Visualization (Nut Assembly).}
    \label{fig:sup1}
\end{figure*}

\begin{figure*}
    \centering
     \includegraphics[width=0.85\linewidth]{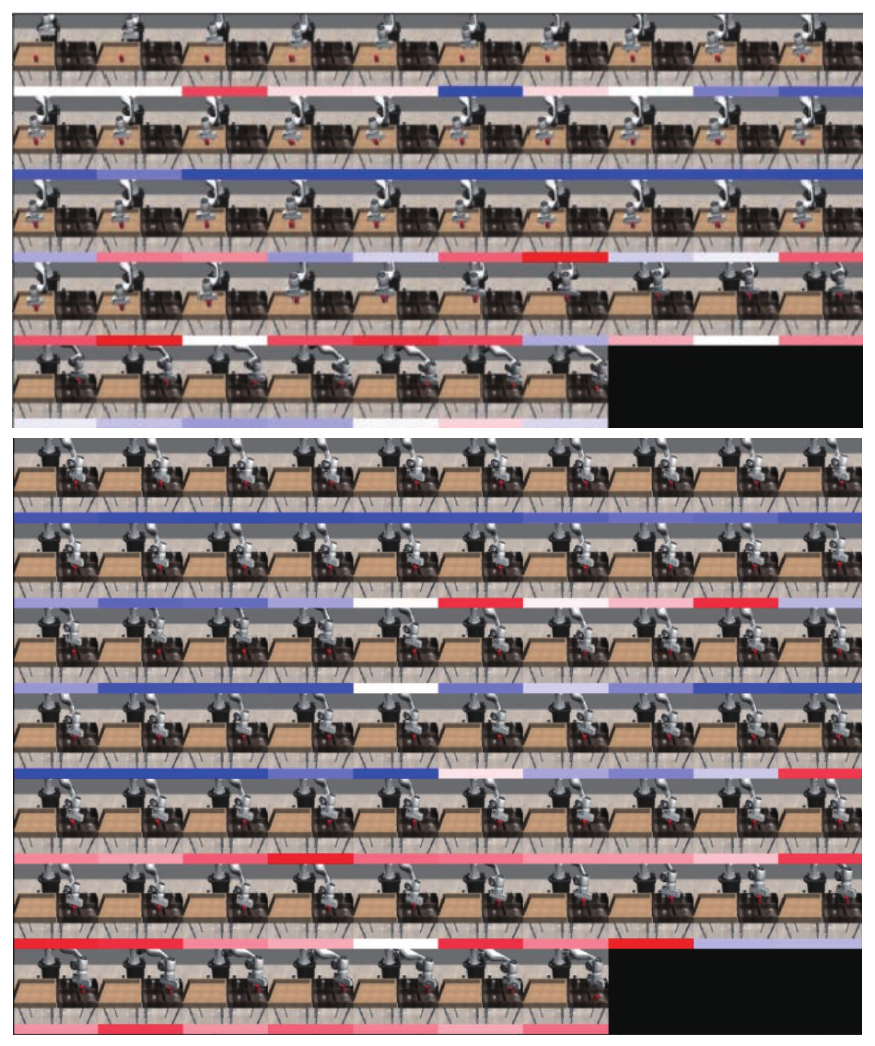}
    \caption{More Weight Visualization (Can PickAndPlace).}
    \label{fig:sup2}
\end{figure*}

We visualize more weight assignment plots in Figure~\ref{fig:sup3},  Figure~\ref{fig:sup4}, Figure~\ref{fig:sup1} and Figure~\ref{fig:sup2}.

